\documentclass[10pt,twocolumn,letterpaper,dvipsnames,table]{article}

\usepackage[pagenumbers]{cvpr} 
\usepackage{nicefrac}       
\usepackage{microtype}      
\usepackage{graphicx}
\usepackage{caption}
\usepackage{algpseudocode}
\usepackage{xspace}
\usepackage[dvipsnames,table]{xcolor}
\definecolor{navyblue}{RGB}{0,0,128}
\usepackage{wrapfig}
\usepackage{ctable}
\usepackage{multirow}
\usepackage{varwidth}
\usepackage{adjustbox}
\usepackage{arydshln}
\usepackage{physics}
\usepackage{hhline}
\usepackage{afterpage}
\usepackage{cuted}
\usepackage{makecell}
\usepackage{float}
\usepackage{cuted}
\usepackage[T1]{fontenc}
\usepackage{lmodern}

\newcommand{\ours}{WOD-E2E}
\newcommand{\ourslong}{Waymo Open Dataset for End-to-End Driving}

\usepackage{stfloats}

\definecolor{custombox}{HTML}{ffcd55}
\definecolor{custombox2}{HTML}{00e89d}
\definecolor{vsaoutcolor}{HTML}{f4af91} 
\definecolor{scoutcolor}{HTML}{0078ff}
\definecolor{cvprblue}{rgb}{0.21,0.49,0.74}

\definecolor{oai-green-050}{HTML}{F4FFF4}
\definecolor{oai-green-100}{HTML}{E9FFE8}
\definecolor{oai-green-200}{HTML}{D9FFD8}
\definecolor{oai-green-300}{HTML}{C9FFC7}
\definecolor{oai-green-400}{HTML}{A6FFA3}
\definecolor{oai-green-500}{HTML}{7CF178}
\definecolor{oai-green-600}{HTML}{51DA4C}
\definecolor{oai-green-700}{HTML}{3FA93B}
\definecolor{oai-green-800}{HTML}{2D712A}
\definecolor{oai-green-900}{HTML}{193718}

\definecolor{oai-gray-000}{HTML}{FFFFFF}
\definecolor{oai-gray-100}{HTML}{FAFAFA}
\definecolor{oai-gray-200}{HTML}{F5F5F5}
\definecolor{oai-gray-300}{HTML}{E5E5E5}
\definecolor{oai-gray-400}{HTML}{FFB7A4}
\definecolor{oai-gray-500}{HTML}{CDCDCD}
\definecolor{oai-gray-600}{HTML}{A8A8A8}
\definecolor{oai-gray-700}{HTML}{747474}
\definecolor{oai-gray-800}{HTML}{393939}
\definecolor{oai-gray-900}{HTML}{000000}

\usepackage[pagebackref,breaklinks,colorlinks,allcolors=cvprblue]{hyperref}

\def\paperID{6662} 
\def\confName{CVPR}
\def\confYear{2025}

\title{\ours: \ourslong\\ in Challenging Long-tail Scenarios}

\author{
  Runsheng Xu\thanks{Equal contributions.} \thanks{Contact emails: Runsheng Xu <\textit{runshengxu@waymo.com}>, Jyh-Jing Hwang <\textit{jyhh@waymo.com}>.}  ,
  Hubert Lin\footnotemark[1] ,
  Wonseok Jeon ,
  Hao Feng ,
  Yuliang Zou  ,
  Liting Sun \\  
  John Gorman ,
  Ekaterina Tolstaya ,
  Sarah Tang ,
  Brandyn White ,
  Ben Sapp \\ 
  Mingxing Tan , 
  Jyh-Jing Hwang\footnotemark[2] ,
  Dragomir Anguelov
  \\ 
  ~ \\
  \textbf{Waymo LLC}
}

\begin{document}

\maketitle

\begin{abstract}
Vision-based end-to-end (E2E) driving has garnered significant interest in the research community due to its scalability and synergy with multimodal large language models (MLLMs). However, current E2E driving benchmarks primarily feature nominal scenarios, failing to adequately test the true potential of these systems. Furthermore, existing open-loop evaluation metrics often fall short in capturing the multi-modal nature of driving or effectively evaluating performance in long-tail scenarios. 
To address these gaps, we introduce the \ourslong\ (\ours). \ours\ contains 4,021 driving segments (approximately 12 hours), specifically curated for challenging long-tail scenarios that  that are rare in daily life with an occurring frequency of less than 0.03\%. Concretely, each segment in \ours\ includes the high-level routing information, ego states, and 360-degree camera views from 8 surrounding cameras.

To evaluate the E2E driving performance on these long-tail situations, we propose a novel open-loop evaluation metric: Rater Feedback Score (RFS). Unlike conventional metrics that measure the distance between predicted way points and the logs, RFS measures how closely the predicted trajectory matches rater-annotated trajectory preference labels.  We have released rater preference labels  for all {\ours} validation set segments, while the held out test set labels have been used for the 2025 \ours\ Challenge.
Through our work, we aim to foster state of the art research into generalizable, robust, and safe end-to-end autonomous driving agents capable of handling complex real-world situations.

\end{abstract}    
\section{Introduction}
\label{sec:intro}

\begin{figure*}[ht!] 

    \centering

    \includegraphics[width=0.8\linewidth]{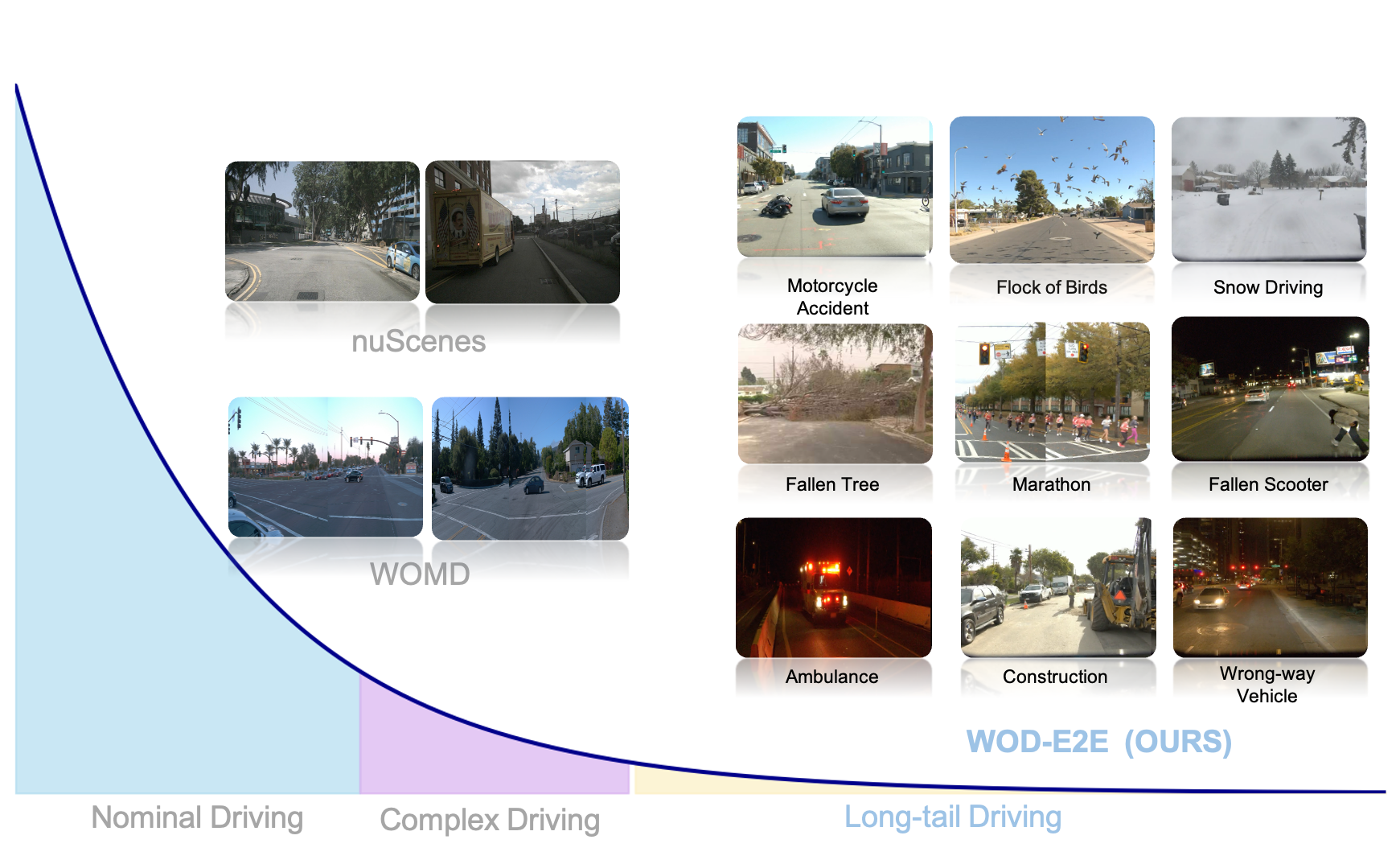}
\hspace{-0.1in}
    \caption{\textbf{Long-tail scenario examples from the Waymo Open Dataset for End-to-End Driving (WOD-E2E).} Unlike existing datasets that are commonly used for E2E Driving benchmarking, \ours\ dataset has more explicit focus on long-tail scenarios. Our analysis in Section~\ref{sec:mining} shows that \ours\ captures the long-tail scenarios with a frequency of less than \textbf{0.03\%} in daily driving.
    \vspace{-12pt}}
    \label{fig:teaser}
\end{figure*}

Autonomous driving systems have traditionally followed a modular design approach that decomposes the driving task into distinct sub-tasks such as perception, prediction, and planning ~\cite{yurtsever2020survey,hwang2022cramnet,li2022bevformer,sun2022swformer,xu2025v2xvit2,pan2024clip}. While this modular design offers benefits in terms of interpretability and debugging, the research community has recently shifted its attention to exploring vision-based end-to-end (E2E) architectures~\cite{xing2025openemma,pan2024vlp,wang2025adawm,cui2025vilad,xu2024drivegpt4}. This shift is primarily driven by the inherent scalability of E2E systems, which directly map raw sensor data to driving actions, reducing the underlying system complexity and the need for rater annotations of intermediate concepts~\cite{xie2025s4}. Furthermore, as previous works~\cite{hwang2025emma,tian2024drivevlm} indicate, there is a promise of leveraging multi-modal large language models (MLLMs) and their world knowledge for E2E driving. 

Despite this promise, current real-world E2E driving datasets, such as NAVSIM~\cite{Dauner2024NEURIPS}, WOMD~\cite{ettinger2021large} and CoVLA~\cite{arai2025covla}, predominantly feature \emph{nominal} driving scenarios that do not fully expose systems to the long tail of possible real-world situations. This scarcity of long-tail examples hinders the accurate evaluation of the true potential, robustness, and generalization ability of E2E driving systems.

In this paper, we introduce the newly released  \ourslong\ (\ours), which explicitly focuses on long-tail situations. As shown in Figure~\ref{fig:teaser}, \ours\  features rare real-world scenarios, which occur with a frequency of less than 0.03\%. 
We provide 4,021 challenging driving segments comprising approximately 12 hours in total, where each segment contains 8 surrounding cameras covering a 360-degree field of view, high-level routing information, ego vehicle position history, and 5s of its future trajectory. These driving segments are collected from a mixture of autonomous and manual driving.  Moreover, we observe that previous open-loop metrics often fail to adequately evaluate the driving performance in these long-tail scenarios. The popular Average Distance Error (ADE) or L2 error metric captures only the error between a prediction and a single future ground truth trajectory, despite the driving behavior being inherently multi-modal, where multiple reasonable future trajectories are possible. Predictive metrics, such as PDMS scores~\cite{Dauner2024NEURIPS}, require annotated positions and future trajectories of road agents to calculate collision rates, and thus become impractical in many long-tail scenarios involving novel or hard-to-detect objects (e.g., the flock of birds shown in Figure \ref{fig:teaser}). Furthermore, off-road behaviors typically incur high penalties in PDMS, yet in numerous safety-critical long-tail scenarios, an autonomous vehicle might reasonably deviate partially off-road to avoid an emergency. To address these limitations, \ours\ dataset also includes a subset of human driving preference labels, providing expert ratings on multiple potential trajectories in each example. Leveraging these labels, we propose a novel open-loop evaluation metric, the Rater Feedback Score (RFS), to better evaluate the E2E driving performance in an open-loop setting. 

We conduct rigorous studies with robust baseline models to verify the dataset and RFS. Since the dataset release, we have garnered significant interest from the research community, with numerous methods already submitted and evaluated on our public leaderboard. The diversity of these top-performing methods, employing approaches such as MLLMs~\cite{pal2025poutinevisionlanguagetrajectorypretraining,wang2025hmvlm}, diffusion models~\cite{liao2025diffusiondrive}, and CNN/ViT with GRU/MLP architectures~\cite{ParkSwinTrajectoryTR}, further underscores the utility of the WOD-E2E dataset and its promise to drive further advances in end-to-end autonomous driving research. Our contribution can be summarized as:

\begin{itemize}
    \item We introduce \textbf{WOD-E2E}, a new open dataset focusing on long-tail scenarios for benchmarking end-to-end autonomous driving systems. It contains 4,021 challenging driving segments, totaling approximately 12 hours of data and representing real-world long-tail scenarios occurring with a frequency of less than 0.03\% in daily driving. 
    \item We propose Rater Feedback Score (RFS), a novel and human-aligned open-loop metric. RFS is designed to better assess E2E driving performance in long-tail scenarios, addressing the limitations of traditional open-loop metrics like ADE and PDMS.
    \item We provide detailed comparison and analysis for our baseline E2E model and multiple methods submitted to our public leaderboard, based on this new dataset. The widespread participation validates the dataset's usefulness for facilitating the E2E  driving research.
\end{itemize}

In the remainder of this paper, we first discuss relate works in Section~\ref{sec:related}. In Section~\ref{sec:dataset}, we describe in detail the proposed \ours\ dataset, including overview, quantitative analysis, mining strategy, labeling, and the rater feedback score metric.  Finally, we summarize all the experimental results and detailed analysis in Section~\ref{sec:exp} and conclude the paper in Section~\ref{sec:conclusion}.

\section{Related Works}
\label{sec:related}

\subsection{End-to-end autonomous driving research}
The paradigm of E2E autonomous driving, directly mapping raw sensor inputs to control outputs, continues to be a vibrant area of research, seeking to overcome the complexities in traditional modular pipelines~\cite{hwang2025emma,xie2025s4}. Recent works have significantly advanced the capabilities of E2E systems, particularly through the use of foundation models. Overall, the current methods can be divided into three categories:

\noindent \emph{Bird's-Eye-View (BEV) Based E2E Planner:}
This type of method aims to fuse information from multiple sensors into a single, comprehensive BEV representation, from which both perception and planning tasks can be directly performed. UniAD~\cite{hu2023uniad}  exemplifies this by propagating BEV queries from its perception module to downstream tasks such as tracking, motion forecasting, and occupancy prediction, ultimately enabling end-to-end planning. Similarly, BEV-Planner \cite{li2024bevplanner}  focuses on learning an explicit planning policy directly from BEV features, demonstrating how dense BEV representations can facilitate robust end-to-end control. These approaches move beyond explicit intermediate perception outputs for planning. Overall, these unified BEV-centric methods offer advantages in terms of computational efficiency and coherence by providing a consistent spatial understanding across various driving sub-tasks.

\noindent \emph{Multi-modal Large Language Model Based E2E Planner:}
A prominent trend involves leveraging Multimodal Large Language Models (MLLMs) to imbue E2E driving systems with enhanced reasoning capabilities and world knowledge. DriveGPT4~\cite{xu2024drivegpt4}  utilizes LLMs to both explain vehicle actions and predict control signals in an iterative question-and-answer format. DriveVLM~\cite{tian2024drivevlm} applies chain-of-thought for end-to-end driving, while VLP~\cite{pan2024vlp}  applies the reasoning of MLLMs directly on the Bird's-Eye-View (BEV) space. EMMA~\cite{hwang2025emma} leverages Gemini  to process multiple driving tasks, including planning, 3D detection, and road understanding, within a unified language space. OpenEMMA~\cite{xing2025openemma} and LightEMMA~\cite{qiao2025lightemma}  follow a similar paradigm to build an open-source and lightweight version, respectively. Additionally, S4-Driver~\cite{xie2025s4} proposes to lift the vision tokens from MLLMs to a 3D space. 

\noindent \emph{Diffusion Based E2E Planner:}
 Diffusion models excel at capturing the multi-modal nature of driving actions and generating diverse, plausible trajectories. Notably, DiffusionDrive~\cite{liao2025diffusiondrive} introduces a truncated diffusion policy and efficient cascade decoder for real-time E2E driving. EnDfuser~\cite{wintel2025using} further explores using diffusion ensembles to estimate uncertainty in trajectory planning, leveraging fused camera and LiDAR features to produce distributions of candidate trajectories.

\subsection{End-to-end autonomous driving open dataset}

A multitude of autonomous driving datasets are available today, supporting a diverse range of driving tasks. Notable examples include Kitti~\cite{geiger2013vision}, Argoverse~\cite{chang2019argoverse}, Argoverse 2~\cite{wilson2023argoverse}, WOD-Perception~\cite{sun2020scalability}, and V2V4Real~\cite{xu2023v2v4real}.
While these datasets serve various purposes, a specific subset focuses on end-to-end driving.
Among the most prominent open datasets in this category are nuScenes~\cite{caesar2020nuscenes}, NAVSIM~\cite{Dauner2024NEURIPS}, WOMD~\cite{ettinger2021large}, and CoVLA~\cite{arai2025covla}.

\subsubsection{nuScenes}
nuScenes~\cite{caesar2020nuscenes} is initially developed for perception tasks and features multiple sensor modalities. While recent research~\cite{tian2024drivevlm,hwang2025emma,pan2024vlp} has explored end-to-end driving on this dataset, often using ADE as a primary performance indicator, the core focus of nuScenes remains perception rather than planning. Some studies~\cite{li2024bevplanner,zhai2023rethinking} have observed that even simple extrapolation of historical behavior can yield strong performance without relying on camera images, suggesting that nuScenes may not be ideally suited for complex planning tasks. 

\subsubsection{NAVSIM}
NAVSIM~\cite{Dauner2024NEURIPS} is a compact simulation and benchmarking framework built upon a filtered version of nuPlan~\cite{caesar2021nuplan}. Its core contribution lies in enabling large-scale real-world evaluation through a non-reactive simulator, which effectively bridges the gap between open-loop and closed-loop testing via simulation-based metrics. While NAVSIM has helped to significantly advance end-to-end driving research, its approach presents two major limitations that motivated our work. First, as a simulation framework, it relies on filtering existing datasets rather than providing a \textbf{raw data collection effort} specifically for long-tail events, which may preclude it from capturing the full, nuanced diversity of real-world long-tail scenarios. Second, the PDMS proposed in NAVSIM—which heavily prioritizes ego progress and comfort, along with time-to-collision (TTC)—may prove insufficient for true safety-critical situations. For instance, TTC is challenging to measure with amorphous obstacles like a flock of birds, as depicted in Figure \ref{fig:teaser}, and the metric of comfort should be secondary to safety when the vehicle must perform an emergency maneuver, such as avoiding a falling scooter (Figure \ref{fig:teaser}). Our dataset and evaluation methodology are explicitly designed to overcome these two limitations by providing targeted, diverse, long-tail data and a more safety-focused scoring mechanism.

\subsubsection{WOMD}
The Waymo Open Motion Dataset (WOMD)~\cite{ettinger2021large} is a component of the broader Waymo Open Dataset, with a specific emphasis on motion prediction and behavior research. Although recent works such as MoST~\cite{mu2024most} and S4-Driver~\cite{xie2025s4} conduct end-to-end driving research on it, WOMD is primarily designed for motion prediction and for modeling complex agent interactions, rather than planning. Furthermore, the lack of full camera images (only embeddings are provided) makes it difficult for external researchers to conduct comprehensive E2E research. 

\subsubsection{CoVLA}
CoVLA~\cite{arai2025covla}  provides a large-scale, richly annotated collection of real-world driving scenarios, integrating vision, language, and action modalities. It is designed to enable the training of Vision-Language-Action models that can generate descriptive scene captions and predict vehicle trajectories. While CoVLA's automated captioning aims for diversity and covers a wide range of common driving conditions, the available information does not detail specific mechanisms for over-sampling or synthesizing rare, safety-critical long-tail events beyond general diversity.

\section{WOD-E2E Dataset}
\label{sec:dataset}

\subsection{Dataset Overview}
This dataset contains 4,021 driving segments mined from real driving logs. Each segment is 20-second long and focused on long-tail scenarios. The dataset is partitioned as: 2,037 segments for training, 479 segments for validation, and the rest 1,505 segments for testing.

\subsubsection{Coordinate System}
This dataset employs two primary coordinate systems: vehicle coordinates and sensor frame coordinates.

 \noindent \textbf{Vehicle Coordinates}: The vehicle coordinate system is located at the ego vehicle's center. The x-axis points forward, the y-axis points left, and the z-axis points upward. All trajectory data is referenced to this vehicle coordinate system.
 
 \noindent \textbf{Sensor Frames}: Each sensor frame is related to the vehicle frame by an extrinsic transformation. For cameras, the frame is centered at the lens. The x-axis points out from the lens, the z-axis points upward, and the y/z plane is parallel to the camera's image plane. This is a right-handed coordinate system.

\subsubsection{Camera Data}
This dataset includes images from eight cameras, providing 360-degree coverage around the vehicle: front, front left, front right, side left, side right, rear, rear left, and rear right. The sensor layout configuration is similar to that described in \cite{sun2020scalability}. For each direction, a single JPEG image is provided. Alongside the image data, we supply camera intrinsics and extrinsics, which define the camera's internal parameters and its position relative to the vehicle's center, respectively. These parameters enable the projection of 3D trajectories onto the camera images. Each driving segment includes 10Hz camera video sequences. Training data spans 20 seconds, while testing data covers 12 seconds, with the subsequent 8 seconds of future data hidden for evaluation purposes. 

\subsubsection{Routing Information}

We provide a routing input for the model in the form of a high-level command, following conventional academic benchmarks \cite{hu2023uniad, casas2021mp3}. 

The high-level command is encoded as an enum \texttt{\{GO\_STRAIGHT, GO\_LEFT, GO\_RIGHT\}}. These commands specify expected driving direction at decision points, such as intersections or highway on/off ramps. \texttt{GO\_STRAIGHT} means the vehicle should continue along the current path, while \texttt{GO\_\{LEFT,RIGHT\}} means the vehicle should take a branching path instead. Note that commands do not refer to micro maneuvers, such as lane changes or nudges around objects on the road, and do not provide any speed profile information.

\begin{figure}[ht]
    \centering
    \includegraphics[width=0.6\linewidth]{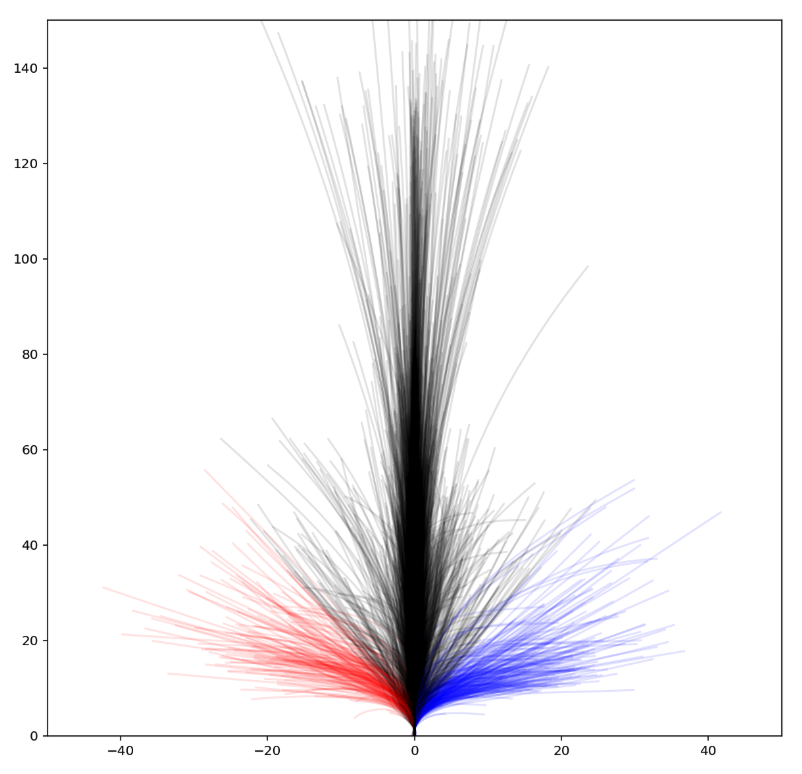}
    \caption{High-level routing input. Ground-truth vehicle trajectories over future \textbf{5s} are shown. Each trajectory is colored \textcolor{red}{red}/black/\textcolor{blue}{blue} corresponding to \textcolor{red}{left}/straight/\textcolor{blue}{right} routing input, derived from \textbf{10s} futures. Units are in meters.}
    \label{fig:hlc}
\end{figure}

We construct high-level commands by comparing the vehicle's \textbf{10s} future driven route against its current position along the route. An illustration is shown in Fig. \ref{fig:hlc}.

\subsubsection{Ego Status}
Each driving segment includes ego vehicle status information, comprising:

 \noindent\textbf{Past Trajectory}: The ego vehicle's past 4-second trajectory, aligned with the current camera timestamp, is provided as waypoints [(x1, y1), (x2, y2),...] at 4Hz frequency. All waypoints are in vehicle coordinates.
 
 \noindent \textbf{Velocity and Acceleration}: The ego vehicle's velocity and acceleration, aligned with its past trajectory, are also provided.
 
 \noindent \textbf{Future Trajectory}: The ego vehicle's future 5-second trajectory from the driving log is provided in the same format and frequency as the past trajectory. This information is available only for the training and validation sets.

\subsubsection{Labels}

 \noindent \textbf{Scenario Cluster}: Each segment is tagged with one of 11 scenario types, which will be explained in detail in the following sections.
 
 \noindent  \textbf{Rater Feedback Labels}: To capture the diversity of acceptable driving decisions during critical events, this dataset includes rater feedback labels. At specific moments within each driving segment, expert labelers rate three distinct 5-second future trajectories on a scale of 0 to 10, where 0 indicates the worst driving and 10 the best. Importantly, we ensure that at least one of the rater-specified trajectories receives a score higher than 6. This label is provided only for the validation set. Details on the creation of these labels will be provided in a subsequent section.

\begin{figure*}[ht!]
    \centering
    \begin{subfigure}[c]{0.48\linewidth} 
        \centering
        \includegraphics[width=\linewidth]{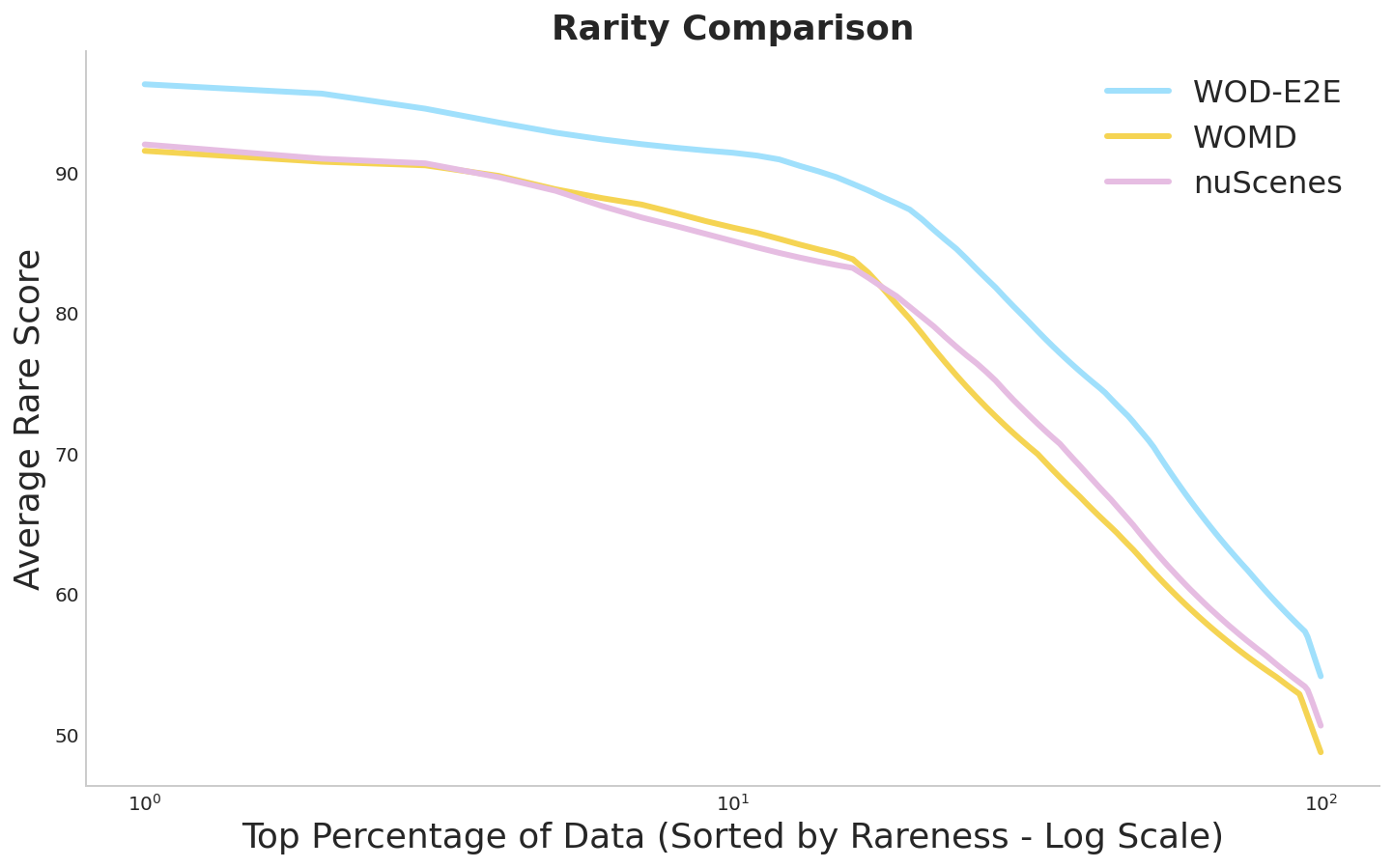}
    \end{subfigure}
    \hfill 
    \begin{subfigure}[c]{0.48\linewidth} 
        \centering
        \includegraphics[width=\linewidth]{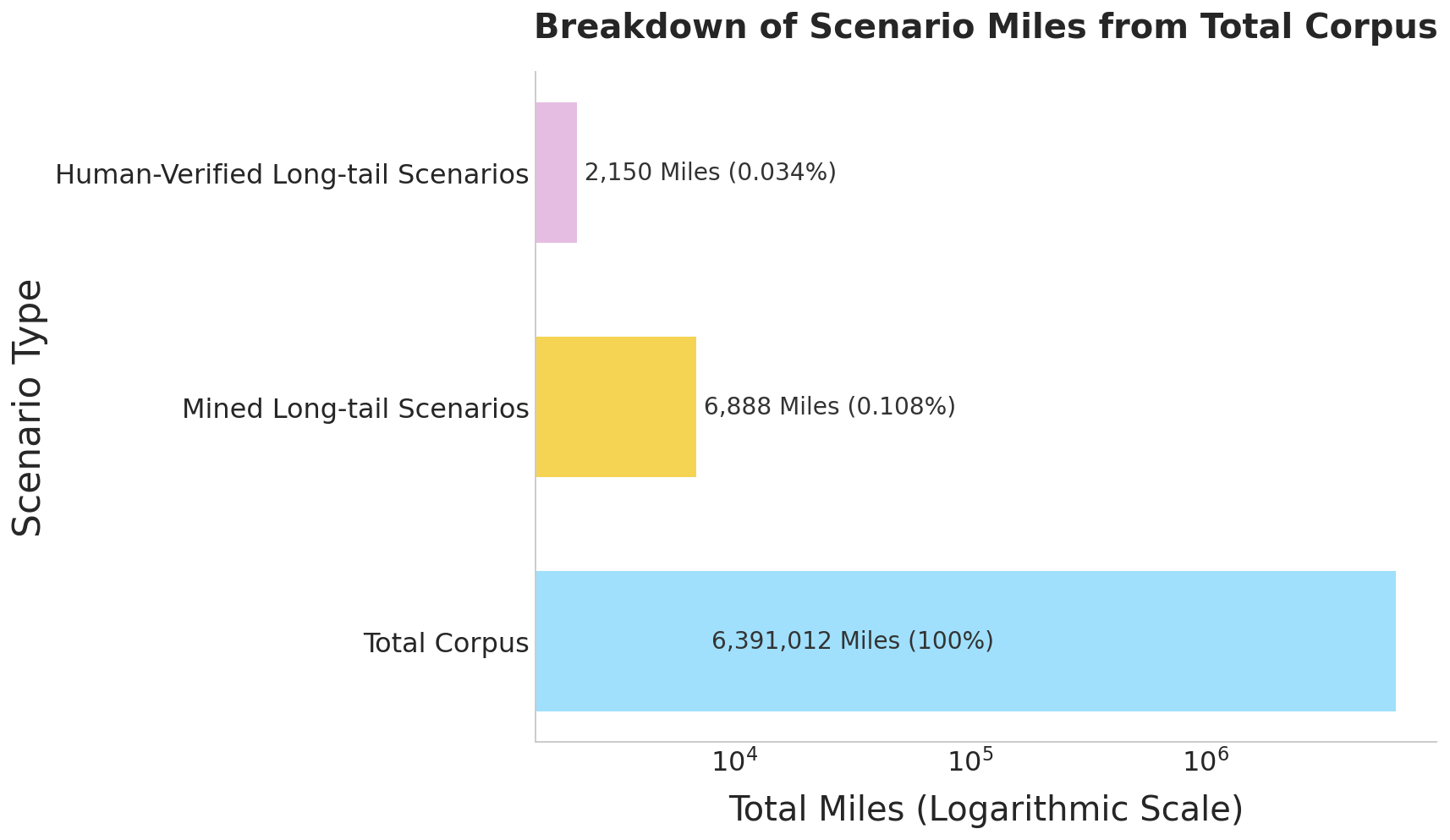}
    \end{subfigure}
    
    \caption{\textbf{Left}: Rarity comparison of driving Datasets. This figure shows the average rarity score for the top percentage of data in each dataset, highlighting the distribution of rare events in \ours. \textbf{Right}: Proportion of mined long-tail scenarios (\textbf{0.03\%}) from the total driving corpus (\textbf{6.4 million miles}).}
    \label{fig:data_comparison}
\end{figure*}

\subsection{Quantitative Rareness Comparison}








In this section, we quantitatively compare the \textbf{rarity} of WOD-E2E against other popular E2E driving datasets. To achieve a standardized, impartial rarity assessment, we utilized a large language model, Gemini 2.5 Pro \cite{comanici2025gemini}, to score the test set of each dataset. The model was provided with the front camera sequences and a detailed scoring prompt outlining four tiers of rarity based on complexity, risk, and long-tail factors. The prompt required the output to be a JSON object containing the \texttt{rarity\_score} that is ranged from 0-100, identified \texttt{rare\_factors}, and a \texttt{reasoning} trace for maximum transparency.

After scoring each scene, we plot the comparative rarity distribution in Figure \ref{fig:data_comparison} (left). This curve is generated by ranking all scenes by their rarity score (high to low) and plotting the average rarity score for all scenes up to that percentage of the dataset.

The figure clearly demonstrates the long-tailed focus of our dataset. We can see that the WOD-E2E curve is significantly higher than all other datasets across all percentage tiles, confirming a higher concentration of long-tail events. Specifically, WOD-E2E maintains a higher average score~(around 93) for the most extreme 10\% of the data, and crucially, its score remains elevated even when considering the full dataset, which indicates the high density of rare scenarios relative to other datasets.

\subsection{Long-tail Data Mining}
\label{sec:mining}

    
    

\subsubsection{Mining Strategy}
We have access to a very large database containing diverse, real-world driving logs that span millions of miles. The vast majority of this data, however, consists of nominal scenarios. To effectively extract only the long-tail scenarios, we developed an efficient mining strategy that combines rule-based heuristics and MLLMs. Firstly, we categorized all driving logs into 11 different categories:

\begin{itemize}
    \item \textbf{Construction:} Scenarios involving construction zones.
    \item \textbf{Intersection:} Scenarios with complex interactions at intersections.
    \item \textbf{Pedestrians:} Scenarios involving interactions with pedestrians.
    \item \textbf{Cyclists:} Scenarios involving interactions with cyclists.
    \item \textbf{Multi-Lane Maneuvers:} Scenarios where the ego vehicle is required to change lanes on multi-lane roads.
    \item \textbf{Single-Lane Maneuvers:} Scenarios where the ego vehicle is required to take actions on single-lane roads.
    \item \textbf{Cut-ins:} Scenarios where other on-road agents cut into the ego vehicle's lane.
    \item \textbf{Foreign Object Debris:} Scenarios with rare objects such as animals or furniture.
    \item \textbf{Special Vehicles:} Scenarios involving special vehicles.
    \item \textbf{Spotlight:} Manually selected challenging scenarios.
    \item \textbf{Others:} Scenarios that do not belong to any of the above clusters.
\end{itemize}

The detailed mining criteria for each category are shown in Table \ref{tab:mining_criteria}. These criteria are made possible by the rich auto-labels available in our dataset, including 3D detection, mapping, tracking, and prediction, which provide the necessary heuristics for our mining process.

\begin{figure*}[ht]
    \centering
    \begin{minipage}[c]{0.48\textwidth}
        \centering
        \begin{subfigure}{\linewidth}
            \centering
            \includegraphics[width=\linewidth]{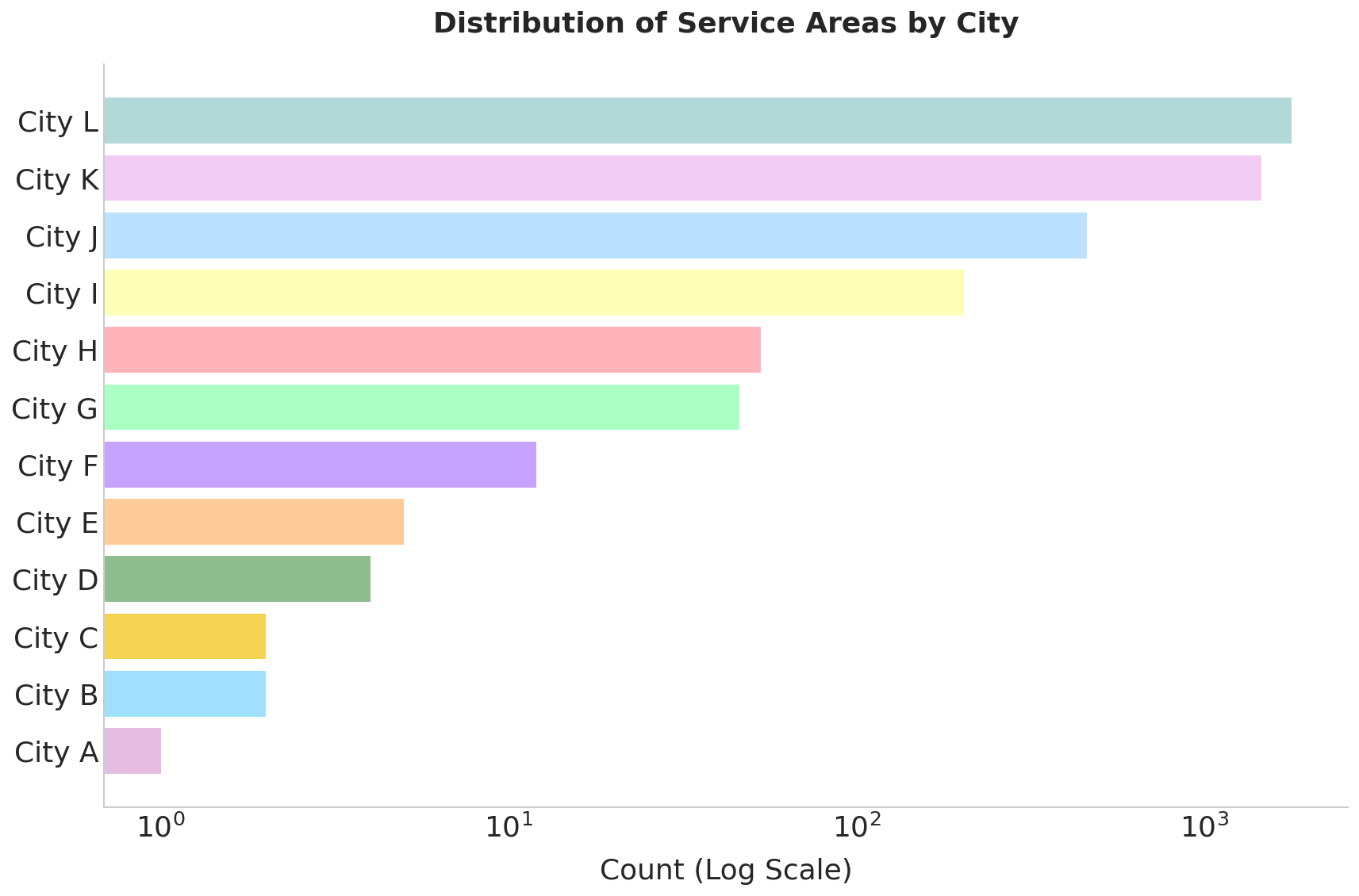}
        \end{subfigure}
        
        \vspace{10pt} 
        
        \begin{subfigure}{\linewidth}
            \centering
            \includegraphics[width=\linewidth]{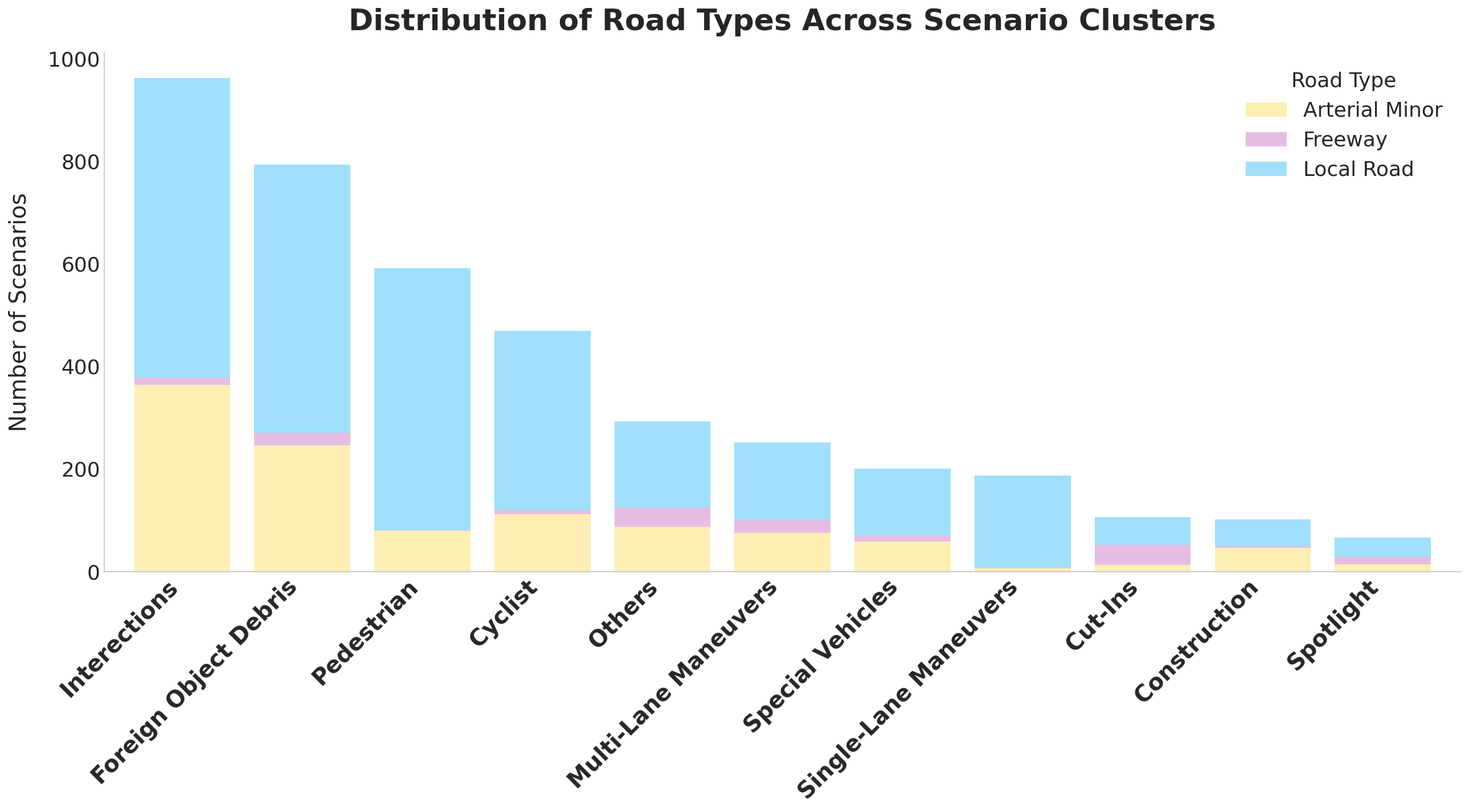}
        \end{subfigure}
    \end{minipage}
    \hfill 
    \begin{minipage}[c]{0.48\textwidth}
        \centering
        \includegraphics[width=\linewidth]{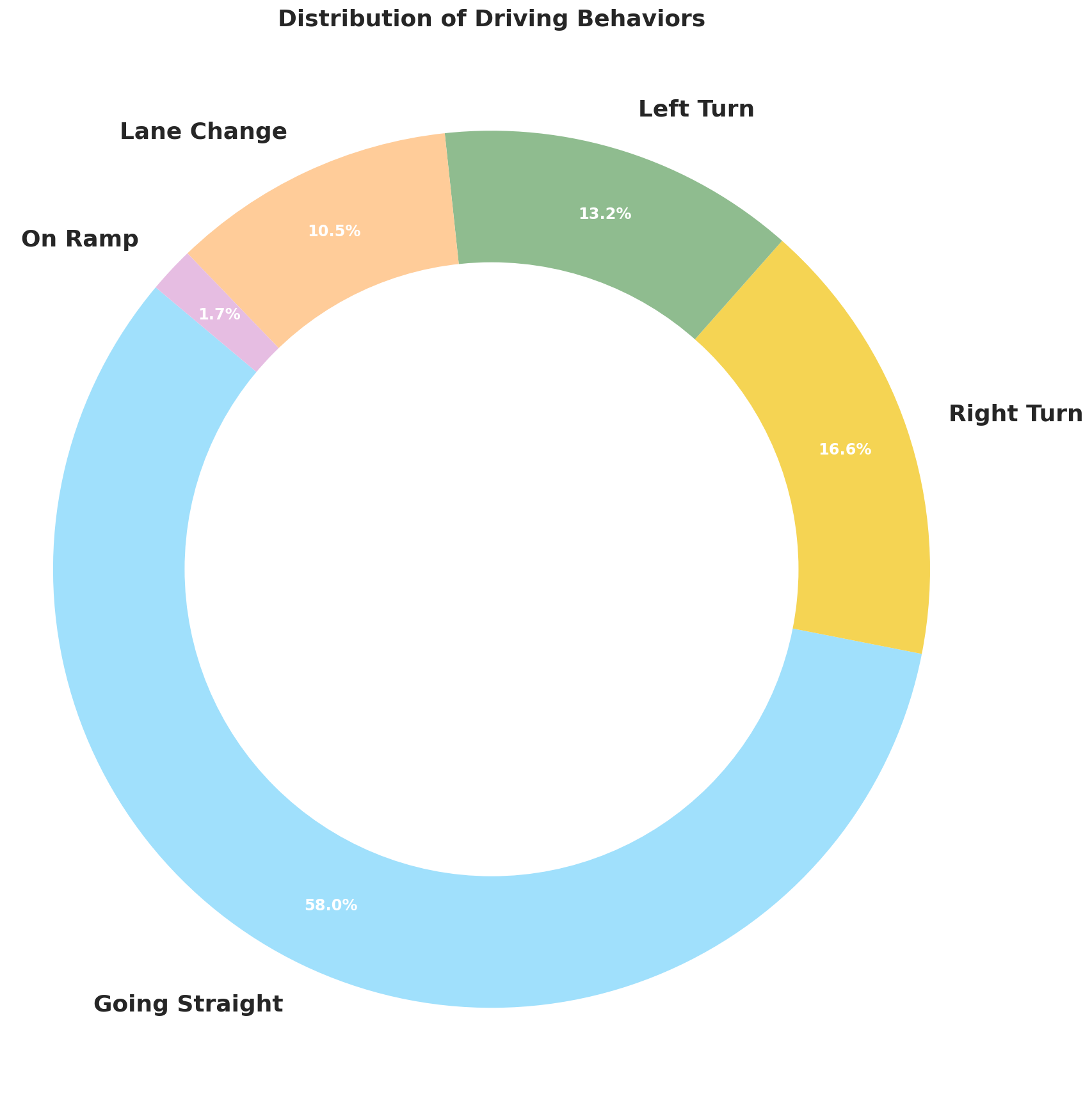}
    \end{minipage}
    
\caption{Comprehensive data distribution analysis. This figure illustrates the key characteristics of the \ours\ dataset across three critical dimensions. \textbf{Top Left}: Distribution of service areas by city. \textbf{Bottom Left}: Distribution of scenario clusters and their breakdowns by road type. \textbf{Right}: Distribution of driving behaviors. }
    \label{fig:combined_distributions}
\end{figure*}

\begin{table*}[ht]
\centering
\caption{Mining criteria for each long-tail scenario category.}
\label{tab:mining_criteria}
\small
\begin{tabularx}{\linewidth}{|*{2}{X|}}
\hline
\rowcolor[HTML]{EFEFEF} 
\textbf{Construction} & \textbf{Intersection} \\
\hline
\begin{itemize}[noitemsep, nolistsep]
    \item Driving route changes due to road closures from a construction zone.
    \item Uniformed pedestrians directing traffic.
    \item Abnormal road surface conditions due to construction.
\end{itemize} &
\begin{itemize}[noitemsep, nolistsep]
    \item Unprotected maneuvers with limited visibility or heavy traffic interactions.
    \item Complex interactions at stop sign intersections.
    \item Interactions with other traffic-violating agents at traffic light intersections.
    \item Interactions with rails and cable cars at intersections.
\end{itemize} \\
\hline
\rowcolor[HTML]{EFEFEF} 
\textbf{Pedestrians} & \textbf{Cyclists} \\
\hline
\begin{itemize}[noitemsep, nolistsep]
    \item Pedestrians crossing with low visibility due to occlusion or weather.
    \item Emergent behavior required to avoid collisions with pedestrians exhibiting unexpected behaviors.
    \item Pedestrians performing unsafe maneuvers specific to the autonomous vehicle.
\end{itemize} &
\begin{itemize}[noitemsep, nolistsep]
    \item Cyclists losing control nearby.
    \item Interactions with a group of cyclists.
\end{itemize} \\
\hline
\rowcolor[HTML]{EFEFEF} 
\textbf{Cut-ins} & \textbf{Foreign Object Debris} \\
\hline
\begin{itemize}[noitemsep, nolistsep]
    \item Oncoming agent cuts across the ego vehicle's trajectory.
    \item An agent in a neighboring lane cuts across the ego vehicle's lane aggressively.
\end{itemize} &
\begin{itemize}[noitemsep, nolistsep]
    \item Interactions with animals on road
        \item Debris that can causes damage on the ADV's path, such as large box, glass debris, and metal debris
        \item Abnormal road condition, such as flooded road, fire on the roadside,severely and degraded road.
\end{itemize} \\
\hline
\rowcolor[HTML]{EFEFEF} 
\textbf{Multi-lane Maneuvers} & \textbf{Single Lane Maneuvers}  \\
\hline
\begin{itemize}[noitemsep, nolistsep]
    \item Nudge maneuvers to overtake blocked agents in the current lane
    \item Lane merging maneuvers on freeway
    \item Other agents in the other lane get too close to ADV that could cause hazards
\end{itemize} &
\begin{itemize}[noitemsep, nolistsep]
    \item Overtake maneuvers in narrow single lane roads
    \item Interactions with open-door vehicle in a narrow single lane road
\end{itemize} \\
\hline
\rowcolor[HTML]{EFEFEF} 
\textbf{Special Vehicles} & \textbf{Spotlight}  \\
\hline
\begin{itemize}[noitemsep, nolistsep]
    \item Emergency vehicles blocking road due to accidents or construction
    \item Pull-over required due to the emergency vehicles
\end{itemize} &
\begin{itemize}[noitemsep, nolistsep]
    \item Leveraging Gemini to search over the database to find scenarios containing certain long-tail objects
\end{itemize} \\
\hline
\end{tabularx}
\end{table*}

\subsubsection{Case Study}
To validate the effectiveness of our mining strategy, we conducted a case study on a recent set of driving logs that includes a total of 6,391,012 miles. After applying our automated mining strategy, we found that only 6,888 miles (0.1\%) of the data fit our criteria for long-tail scenarios. This initial result shows that our strategy is highly effective at isolating rare, challenging events from a massive volume of nominal driving data, as demonstrated in the right figure of Figure \ref{fig:data_comparison}.

Moreover, to ensure the highest quality of our dataset, we perform a subsequent round of human filtering. This manual review process, which has a conversion rate of 30\%, further refined the mined data by removing non-long-tail scenarios. This final filtering step reduced the overall portion of long-tail scenarios to an even rarer 0.03\%, highlighting the signficant infrequency of these critical events in real-world driving.

\subsubsection{Data Analysis}

As shown in Figure \ref{fig:combined_distributions}, we analyze the dataset's distributions across three key dimensions: city locations, scenario clusters, and driving behaviors.

\noindent\textbf{City Distribution.} The top-left subfigure shows the geographical distribution of the dataset across different cities. For confidential reasons, all city names have been anonymized. The data is predominantly sourced from cities L, K, and J, and the remaining cities, which are only present in the test set, contribute a smaller but more diverse set of scenarios, which is crucial for evaluating model generalization.

\noindent\textbf{Scenario Clusters.} The bottom-left subfigure provides a clear overview of our dataset's composition by problem cluster and road type.
\begin{itemize}
    \item We first analyze the distribution of long-tail scenarios by their problem clusters. The clusters for Intersections, Foreign Object Debris (FOD), and Pedestrians account for the largest share of the dataset. This highlights our focus on a variety of complex and safety-critical events, including intricate interactions at intersections, challenging scenes for the perception module, and high-risk encounters with pedestrians.
    \item Our dataset contains three major road types: Local Road, Arterial Minor, and Freeway. The Freeway road type is most prominent in the Cut-ins cluster, which is a particularly safety-critical event at high speeds. It is also notably present in the Intersections cluster. This is because these scenarios specifically capture interactions at freeway entrances and exits, such as making a right turn to enter an on-ramp.
\end{itemize}

\noindent\textbf{Driving Behavior Distribution.} The right subfigure shows the distribution of driving behaviors. We have a variety of diverse behaviors, including moving straight, lane changes, left turns, right turns, and on-ramp maneuvers. The majority of behavior is moving straight, which includes typical lane-following, but also hard braking and swerving for emergency situations. Turning behaviors at intersections, including left and right turns, make up approximately 30\% of the data, with roughly equal proportions. Additionally, lane changes account for 10.3\% of the scenarios, which usually involve collision or obstacle avoidance. Finally, a small portion of the data (1.7\%) is dedicated to on-ramp behaviors, which are often challenging to tackle due to the interaction of merging vehicles at high speeds.

\subsection{Data Labeling}

The mined data is sent to our data labeling pipeline, which consists of three major steps: critical moment selection, trajectory sampling, and trajectory scoring. 

\begin{figure}[ht]
    \centering
    \includegraphics[width=\linewidth]{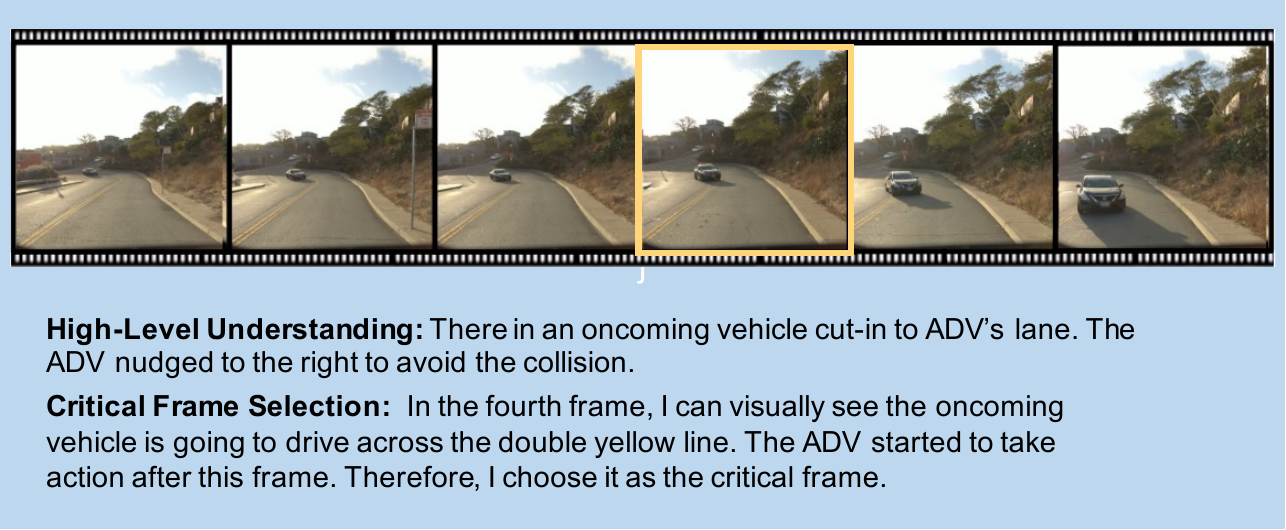}
    \caption{An illustration of how a critical frame is selected.  The human raters first scan through the video for high-level understanding, and then select the critical frame, which is the earliest moment when a critical event is visually apparent in the camera images. Finally, the raters also document the rationales for the critical frame selection.}
    \label{fig:critical_moment}
\end{figure}

\subsubsection{Critical Moment Selection}
The critical moment is defined as the specific frame where a critical event emerges, requiring the vehicle to make an important driving decision. These decisions can include actions like slowing down, nudging, or giving way to other vehicles in the scene. An example can be found in Figure~\ref{fig:critical_moment}.

We instruct our labelers to follow a three-step process for selecting the precise moment:
\begin{enumerate}
\item \textbf{High-level Understanding}: Labelers must first scan the entire video to understand the critical event within the segment and identify the correct driving decision to be made.
\item \textbf{Moment Selection Based on Visual Cues}: Labelers must then find the earliest moment where the critical event is visually apparent in the camera feed. They are instructed to select the frame where the autonomous vehicle has already started taking action to avoid reaction bias introduced by the history motion information. This is typically the frame where the target behavior is most clearly exhibited, such as the initial moment of a lane change or the point of start braking.
\item \textbf{Reasoning Documentation}: The final step involves briefly documenting the rationale for selecting the specific frame. This documentation ensures consistency and provides valuable feedback for model training and analysis.
\end{enumerate}

\subsubsection{Trajectory Sampling}
Trajectory sampling is the process of generating a diverse set of possible motion plans for later human review and selection in a specific driving scenario. Our approach utilizes an existing machine learning model, such as Wayformer~\cite{nayakanti2023wayformer}, to produce an initial set of up to 64 diverse trajectories for a given critical moment. These trajectories are generated using various inputs, including perception detections, mapping elements, and predicted behaviors of other road agents.

 Our trajectory selection process employs a two-step approach that leverages both automated filtering and human-guided selection to identify the most representative motion plans for rating. Initially, the generated trajectories are automatically sorted into different ``buckets'' based on driving decisions, such as velocity, acceleration, and lane changes. From these buckets, we sample a set of diverse candidates (usually fewer than 12). This sampling typically involves selecting the leftmost, middle, and rightmost trajectories to capture a spectrum of lateral movements. This small set of diverse trajectories is then passed to human labelers. The labelers' task is to select three trajectories from these candidates for final ranking and reasoning, ensuring the labeled data includes the optimal path alongside plausible alternative and suboptimal behaviors.

\subsubsection{Trajectory Scoring}

The sampled trajectory candidates, along with the selected critical scenario, are sent to trained human raters under a rigorous manual grading process.

\begin{enumerate}
\item \textbf{Scenario Representation}: The selected long-tail scenarios are represented within a visualization tool to ensure effective and precise labeling. Each scenario is $\mathbf{20}$ seconds long and includes comprehensive data, such as mapping elements, camera images, and annotations for all on-road agents. Candidate trajectories are also plotted directly in this environment. Labelers can easily navigate different timestamps to precisely visualize how each candidate trajectory interacts with the logged future behavior of other road agents or static map elements. This capability is crucial for informed decision-making.

\item \textbf{Trajectory Selection and Grading Criteria}: Within a selected scenario, raters first select three diverse trajectories from the available candidates. This selection must include at least one trajectory that is considered optimal or appropriate behavior, while the other two should represent different behavioral modes that may be sub-optimal. The labelers then rate these three trajectories based on five distinct dimensions:

    \noindent \textbf{Safety:} Whether the trajectory results in collisions, near-misses, or other unsafe conditions.
    
    \noindent \textbf{Legality:} Whether the trajectory complies with all traffic laws and regulations, including proper behavior around emergency vehicles.
    
    \noindent\textbf{Reaction Time:} Whether the autonomous vehicle's actions within the trajectory are timely in response to unfolding events.
    
    \noindent\textbf{Braking Necessity:} Whether the trajectory includes unnecessary, sudden, or overly conservative braking.
    
    \noindent\textbf{Efficiency:} Whether the trajectory demonstrates efficient progress, avoiding unnecessary lane changes, hesitations, or over-reactions to distant or irrelevant agents.

\item \textbf{Scoring Mechanism}: Trajectories are scored on a scale from 0 (worst) to 10 (perfect). Each trajectory is initialized with a base score of $\mathbf{10}$ \textbf{points}. Points are then deducted based on violations of the grading criteria:
\begin{itemize}
    \item Major infractions: A deduction of $\mathbf{2}$ \textbf{points} is applied for violations related to safety, reaction time, or legal violations.
    \item Minor infractions: A deduction of $\mathbf{1}$ \textbf{point} is applied for violations related to braking necessity or efficiency.
\end{itemize}
These penalties are cumulative. In cases where a trajectory exhibits multiple concurrent violations, raters may apply additional discretionary deductions to reflect the severity of the combined faults, ensuring the final score accurately reflects the trajectory's overall quality.
\end{enumerate}

\begin{figure}
    \centering
    \includegraphics[width=\linewidth]{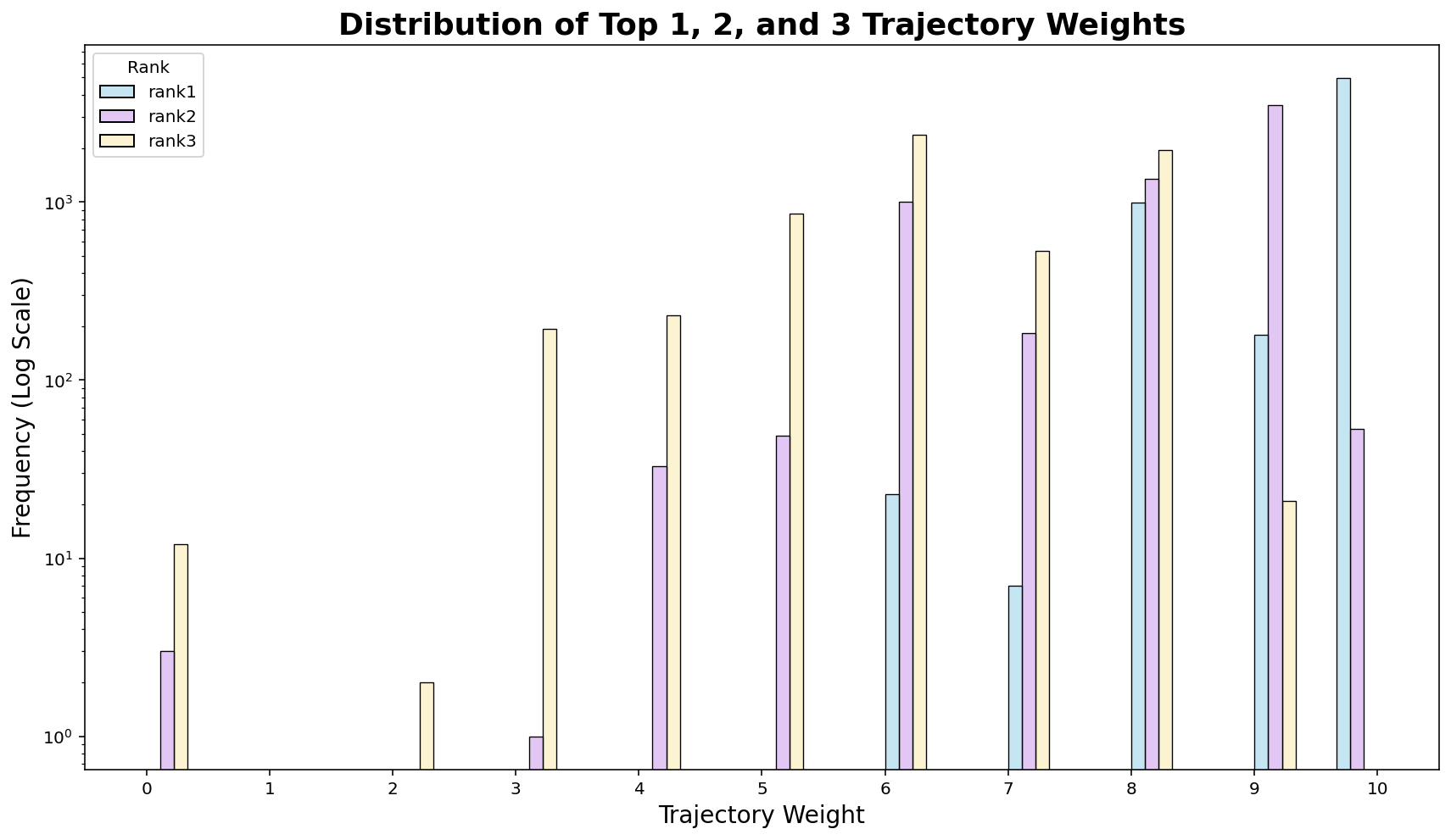}
    \caption{Human rating distribution among 3 candidate trajectories in the \ours\ dataset. This plot revelas a deliberate separation of trajectory quality.
    By including plausible sub-optimal and unsafe alternatives (Rank 3) alongside the optimal path (Rank 1), the label distribution provides essential boundaries for estimating robust end-to-end driving models.
    }
    \label{fig:label_distribution}
\end{figure}

The distribution of final human ratings for the top three trajectories is visualized in Figure $\ref{fig:label_distribution}$. This plot clearly demonstrates a deliberate separation of trajectory quality: The Rank 1 trajectory shows a strong bias towards optimal behavior, with its lowest observed score being 6, which is the minimum score required to regard a trajectory as safe and feasible. In contrast, the Rank 2 and Rank 3 trajectories span a much wider range, with a significant amount of data, particularly for Rank 3, falling below a score of 6. This diverse scoring range successfully captures the desired multi-modality in driving behavior. By including plausible sub-optimal and unsafe alternatives alongside the optimal path, the label distribution provides essential boundaries for estimating robust end-to-end driving models.

\subsection{Rater Feedback Score}
The Rater Feedback Score (RFS) is a metric designed to evaluate the quality of a model's predicted trajectory with the reference of multiple human-annotated trajectories. The WOD-E2E dataset includes 3 reference trajectories generated by human raters, each assigned a score $s_{\mathrm{rater}}$ in $[0, 10]$.

\begin{figure}
    \centering
    \includegraphics[width=0.95\linewidth]{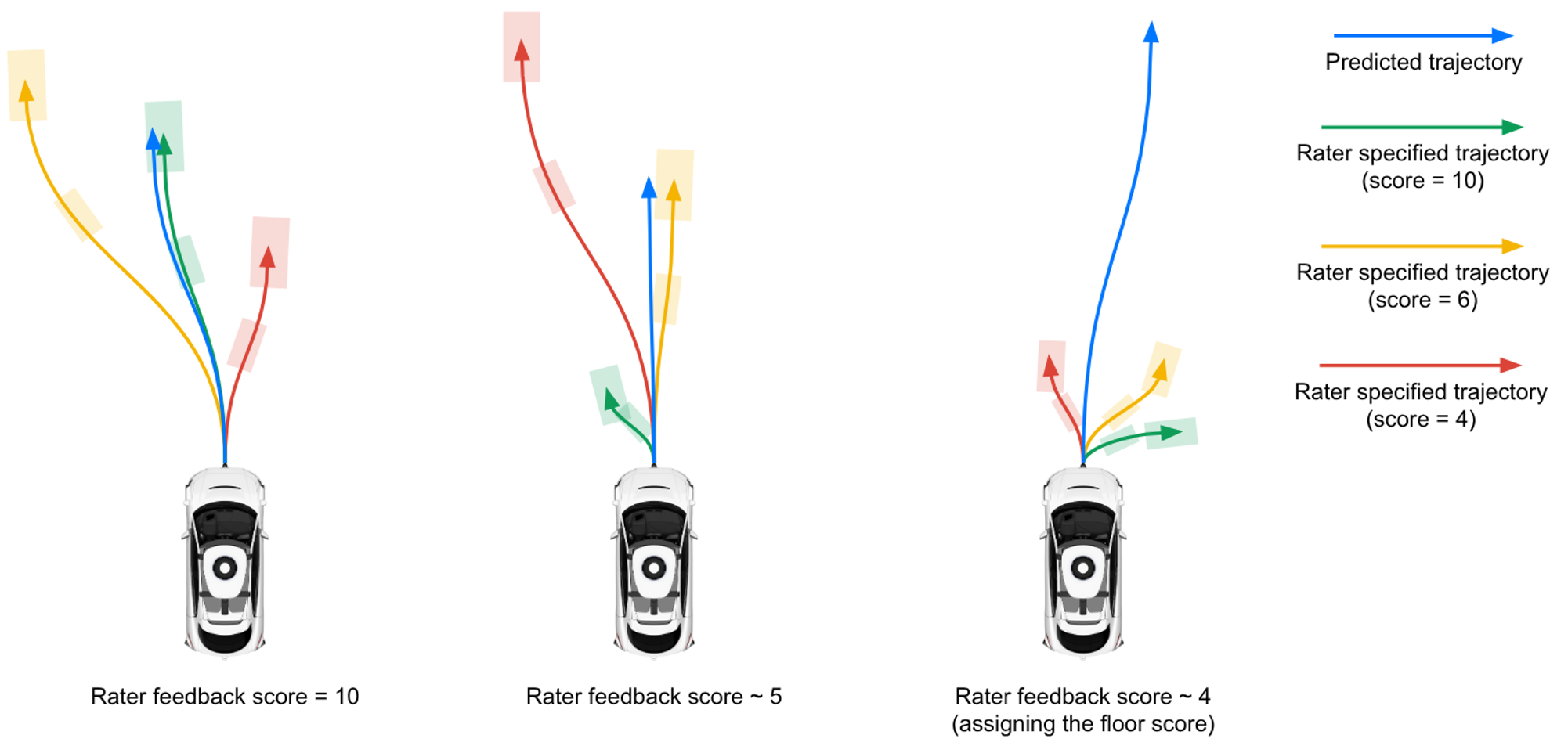}
    \caption{\textbf{Rater Feedback Score  Mechanism.} This figure illustrates how the RFS evaluates a model's Predicted Trajectory (Blue) against three human-rated reference trajectories. The predicted score is based on the highest-rated reference trajectory it aligns with within the defined trust region.}
    \label{fig:rfs_example}
\end{figure}

The RFS is designed to see how much the model's prediction is aligned with three rated trajectories by considering trust regions, as illustrated in Figure~\ref{fig:rfs_example}.
A \textit{trust region} is defined around each rater trajectory at evaluation times $t$ in $\{3, 5\}$ seconds. This region represents the rectangular space within specified longitudinal and lateral distance thresholds from the rater trajectory at a given time $t$.

The base thresholds follows WOMD~\cite{ettinger2021large}, and they are set as $\bar{\tau}_{\mathrm{lat}}=1.0, \bar{\tau}_{\mathrm{lng}}=4.0$ at $t=3$ and $\bar{\tau}_{\mathrm{lat}}=1.8, \bar{\tau}_{\mathrm{lng}}=7.2$ at $t=5$, where the longitudinal threshold $\bar{\tau}_{\mathrm{lng}}$ is always set to be 4 times larger than the lateral threshold $\bar{\tau}_{\mathrm{lat}}$. These base thresholds are scaled based on the initial speed $v$ (m/s) of the rater trajectory. The scaling function is a piece-wise linear function of $v$:
\begin{align*}
    \mathrm{scale}(v)
    =
    \begin{cases}
    0.5, &v<1.4,\\
    0.5+0.5\times\frac{v-1.4}{11-1.4}, &1.4\le v < 11,\\
    1, &v\ge 11.
    \end{cases}
\end{align*}
The final thresholds at $t=3,5$ are determined by 
\begin{align*}
    \tau_{\mathrm{lng}}=\mathrm{scale}(v)\times\bar{\tau}_{\mathrm{lng}}, 
    \tau_{\mathrm{lat}}=\mathrm{scale}(v)\times\bar{\tau}_{\mathrm{lat}}.
\end{align*}
For distance errors $\Delta_{\mathrm{lng}}$ (longitudinal) and $\Delta_{\mathrm{lat}}$ (lateral) and the final thresholds, the score from each rater feedback trajectory is defined by 
\begin{align*}
    s_{\mathrm{rater}}\times 0.1^{\max\left\{\max\left\{\frac{\Delta_{\mathrm{lng}}}{\tau_{\mathrm{lng}}}, \frac{\Delta_{\mathrm{lat}}}{\tau_{\mathrm{lat}}}\right\}-1, 0\right\}}.
\end{align*}
Intuitively, we assign either the flat score $s_{\mathrm{rater}}$, if a predicted trajectory is within the trust region, or the score exponentially decayed from $s_{\mathrm{rater}}$. Then, the final score is determined by choosing the maximum score over all rater specified trajectories, followed by averaging over $t=3,5$ and flooring with $4$. 

\section{Experimental Results}
\label{sec:exp}

\begin{figure*}[ht!]
    \captionsetup{type=table}
    \vspace{-0.4cm}
    \centering
    \begin{minipage}{0.63\textwidth} 
    \centering
    \fontsize{4 pt}{4.4pt}\selectfont
    \setlength\tabcolsep{3pt} 
    \renewcommand{\arraystretch}{1.2} 
    \scalebox{1.57}{
    \begin{tabular}{r|cc|cccc}
    & & &
    \rotatebox{75}{\# of Dataset} &
    \rotatebox{75}{Training Schema} &
    \rotatebox{75}{Model Arch.} &
    \rotatebox{75}{Model Param.} \\
    Methods & RFS$\uparrow$ & ADE$\downarrow$ & \multicolumn{2}{c}{\cellcolor{orange!10}Training Strategy} & \multicolumn{2}{c}{\cellcolor{yellow!10}Model Setting} \\
    \hline
    \rowcolor{navyblue!5}
    \multicolumn{1}{l|}{\textcolor{black}{\textit{MLP-based}}} & & & & & &  \\
    Swin-Trajectory & 7.543  & 2.814 & 1 & SFT & Swin Transformer & 36M \\
    \hline
    \rowcolor{navyblue!5}
    \multicolumn{1}{l|}{\textcolor{black}{\textit{Diffusion-based}}}  & & & & & & \\
    DiffusionLTF & \cellcolor{oai-green-200}{7.717} & 2.977 & 4 
    & SFT &  DiffusionDrive & 60M  \\
    UniPlan & 7.779 & 2.986 & 2 & SFT
    & DiffusionDrive & 60M \\
    \hline
    \rowcolor{navyblue!5}
    \multicolumn{1}{l|}{\textcolor{black}{\textit{MLLM-based}}} & & & & & & \\
    Baseline & \cellcolor{oai-gray-500}{7.528} &3.018 & 1 & SFT & Gemini1 Nano & 3B  \\
    AutoVLA & 7.556 & 2.958 & 3& SFT+RL & Qwen2.5 & 3B  \\
    HMVLM & \cellcolor{oai-green-400}{7.736} & 3.071 & 1 & SFT & Qwen2.5 & 3B \\
    Poutine & \cellcolor{oai-green-600}{7.986} & 2.741 &  2 & SFT+RL & Qwen 2.5 & 3B  \\
    \hline
    \rowcolor{navyblue!5}
    \end{tabular}
    } 
    \end{minipage}
    \hfill
    \begin{minipage}[c]{0.35\textwidth}
        \centering
        \includegraphics[width=\textwidth]{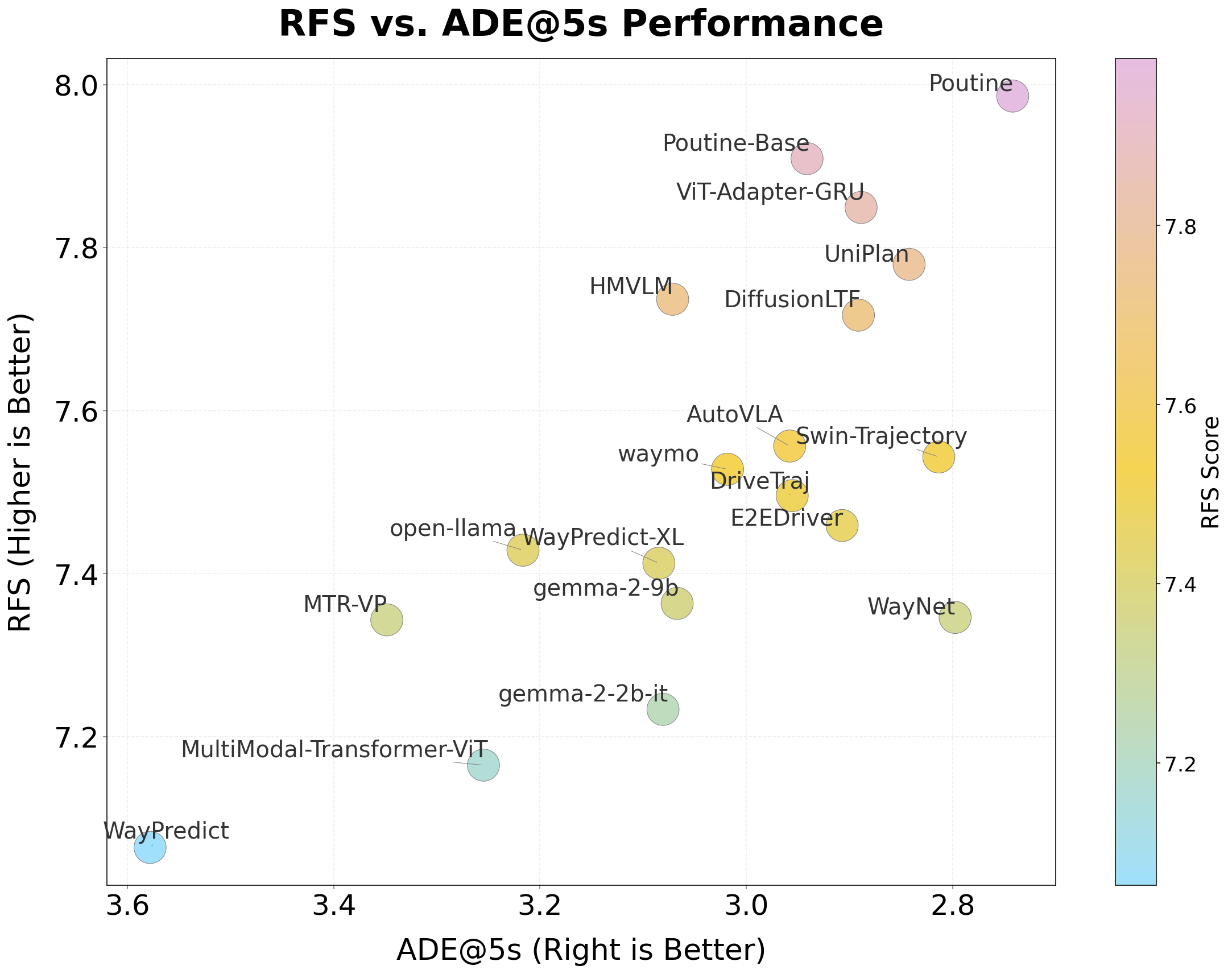} 
        \label{fig:ade_rfs}
    \end{minipage}

    \caption{\ours\ leaderboard submission results. \textbf{Left:} We summarize the results and configurations of selected representative methods among 3 categorical methodology (MLP-based, Diffusion-based, and MLLM-based).  \textbf{Right:} We plot RFS vs ADE using 19 submissions. We only observe a mild positive correlation between RFS and ADE. }
    \label{tab:main_table}
    \vspace{-0.4cm}
\end{figure*}

\subsection{Baseline Model Setup}

\begin{figure}[ht]
    \centering
    \includegraphics[width=\linewidth]{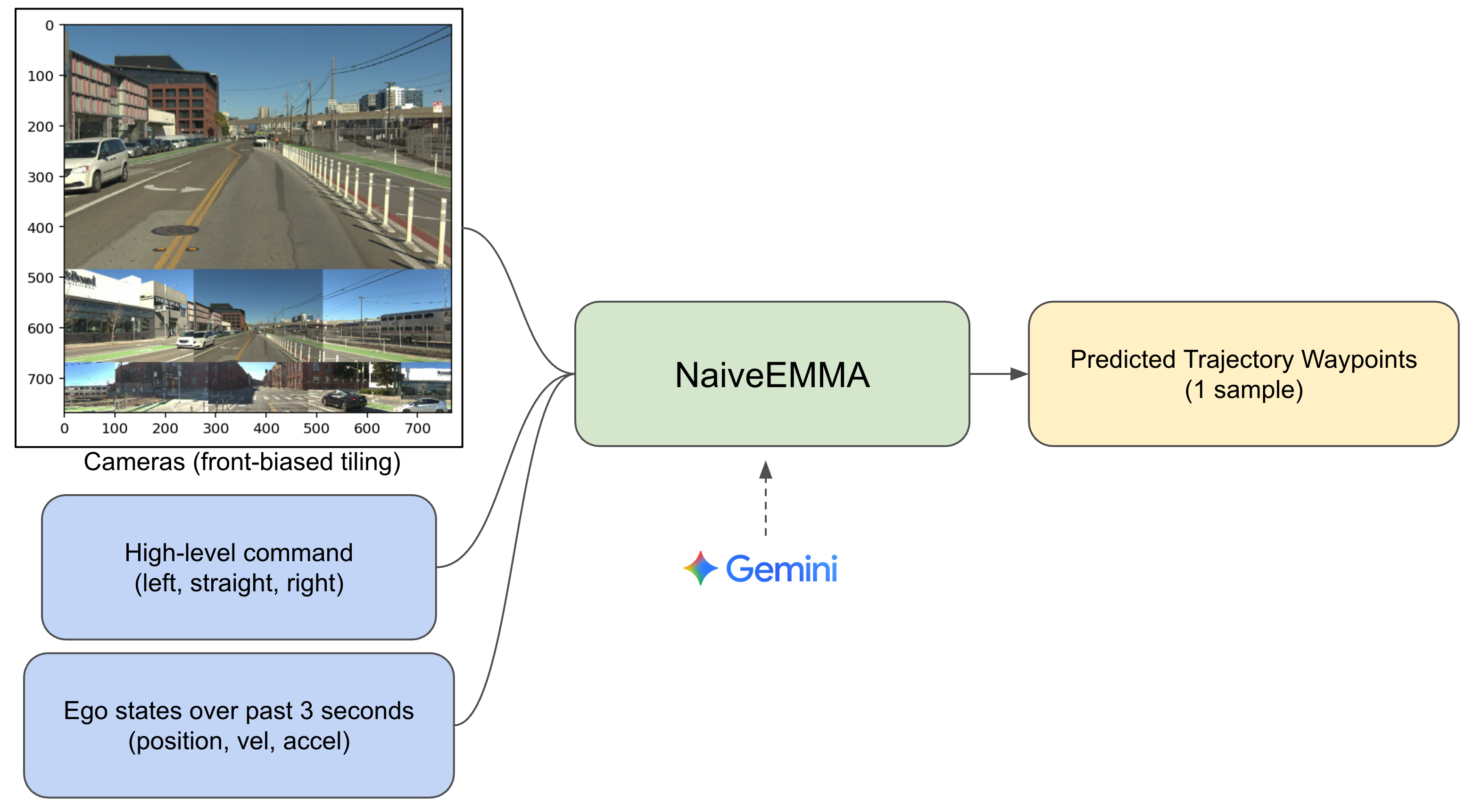}
    \caption{Architecture of NaiveEMMA, which serves as the challenge leaderboard baseline.  NaiveEMMA is a highly simplified version of EMMA~\cite{hwang2025emma}, fine-tuned from Gemini Flash~\cite{comanici2025gemini}. The model takes as input all 8 camera images, 3 seconds of past ego-status history, and the high-level routing input. It then predicts the future trajectory in 5 seconds}
    \label{fig:naive_emma}
\end{figure}

We use a highly simplified version of EMMA \cite{hwang2025emma}, which we call NaiveEMMA, as our baseline model. The architecture of NaiveEMMA is illustrated in Figure \ref{fig:naive_emma}. NaiveEMMA is finetuned directly from Gemini Flash \cite{comanici2025gemini} and has not been trained on any internal driving datasets: it is finetuned exclusively on the released WOD-E2E training split. The model consumes a combined image from all eight cameras at the current timestep, concatenated into a single $768 \times 768$ resolution image. It also takes in 3 seconds of past ego-status history and the high-level routing input. Crucially, it does not use past camera frames. Note that NaiveEMMA omits several advanced components of the original EMMA model, specifically generalist task training mixtures, Chain-of-Thought reasoning, and any test-time scaling methods.

\subsection{RFS Metric Validation}
\subsubsection{Quantitative Validation}

 We train several models based on NaiveEMMA and evaluate RFS on an internal test split. This test split contains long-tailed scenarios similar to the WOD-E2E test split. This experiment controls for several factors that are expected to improve model quality in long-tail settings: exposure to long-tailed scenarios via the WOD-E2E training split, multi-camera inputs to reason about surroundings, and test-time scaling to handle scenario ambiguities. RFS aligns with these intuitions, assigning higher scores to models that utilize more of these features (Table~\ref{fig:rfs_ablate}).


\begin{table}[h!]
\centering
\small
\begin{tabularx}{\linewidth}{|l|X|}
    \multicolumn{1}{l}{\textbf{Model}} & \multicolumn{1}{l}{\textbf{RFS}} \\
    \hline
     Baseline & 7.14\\
    \hline
    + WOD E2E finetuning & 7.22\\
    + multi-camera inputs & 7.30\\
    + test-time scaling (multi sampling) & 7.39 \\
    \hline
\end{tabularx}
\caption{RFS assigns higher scores to models that are better-equipped to handle long-tailed scenarios. Evaluation is performed on an internal test split.}
\label{fig:rfs_ablate}
\end{table}


\begin{figure*}[t]
    \centering
    
    \begin{subfigure}{\linewidth}
        \centering
        \includegraphics[width=0.3\linewidth]{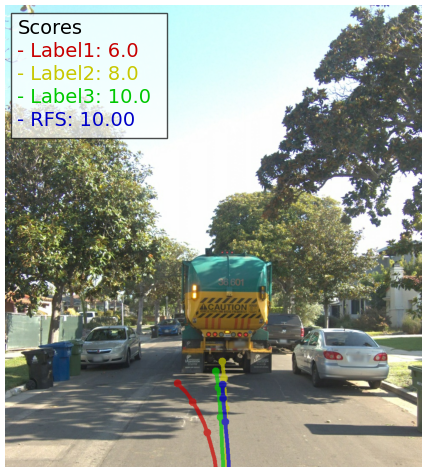}
        \includegraphics[width=0.3\linewidth]{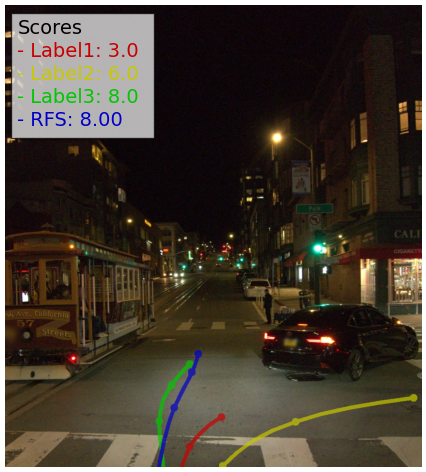}
        \includegraphics[width=0.3\linewidth]{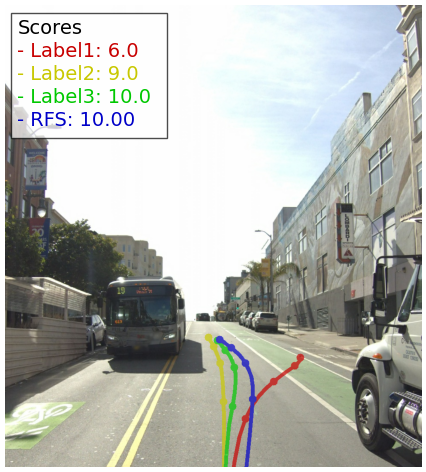}
        \caption{The model predicted future trajectory (\textcolor{blue}{blue}) aligns well with one of the rater specified trajectories. The corresponding flat scores are assigned as the predictions fall within the trust region.
        }
        \label{fig_quant:1}
    \end{subfigure}

    \begin{subfigure}{\linewidth}
        \centering
        \includegraphics[width=0.3\linewidth]{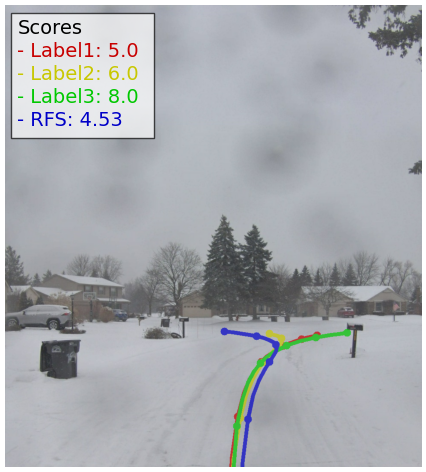}
        \includegraphics[width=0.3\linewidth]{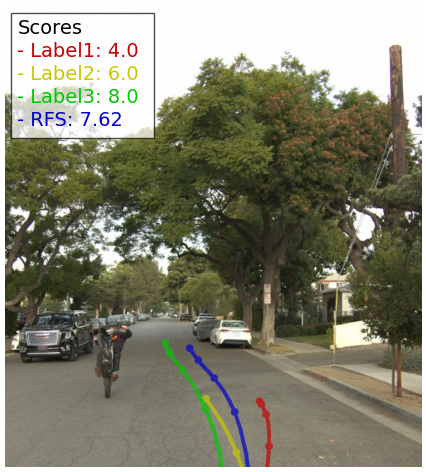}
        \includegraphics[width=0.3\linewidth]{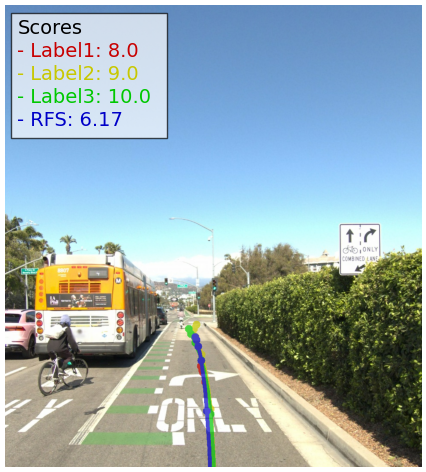}
        \caption{The model predicted future trajectory (\textcolor{blue}{blue}) deviates from rater specified trajectories.  Since the predictions fall outside the trust regions, final scores are exponentially decayed.}
        \label{fig_quant:2}
    \end{subfigure}

    \begin{subfigure}{\linewidth}
        \centering
        \includegraphics[width=0.3\linewidth]{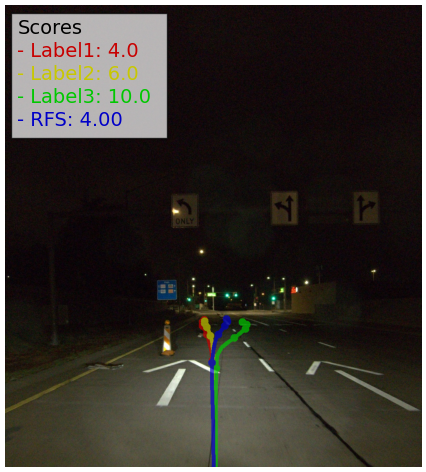}
        \includegraphics[width=0.3\linewidth]{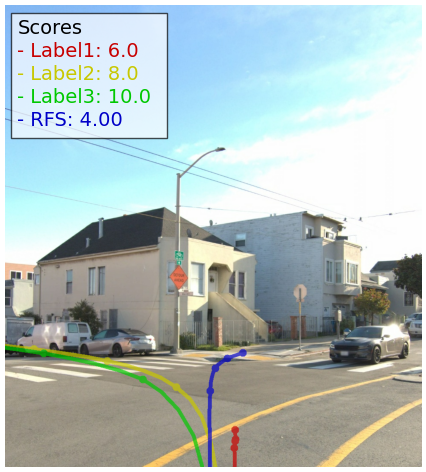}
        \includegraphics[width=0.3\linewidth]{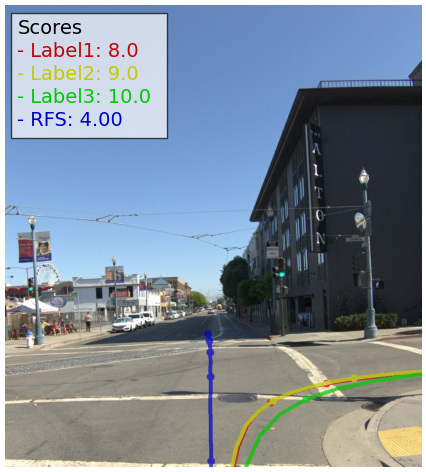}
        \caption{Floored scores (RFS=4) are assigned because predictions are far from any of the rater-specified trajectories.}
        \label{fig_quant:3}
    \end{subfigure}
    
    \caption{Visualization for the RFS metric in 3 different conditions. \textbf{Top:} The model predictions fall within the trust region.  \textbf{Middle:} The model predictions fall slightly outside the trust region.  \textbf{Bottom:} The model predictions are far from any of the rater-specified trajectories.
    }
    \label{fig_quant:0}
    \vspace{0.03cm}
\end{figure*}

\subsubsection{Qualitative Validation}

In this section, we validate the RFS metric through several qualitative examples, as Figure~\ref{fig_quant:0} shows.

\noindent\textbf{Scores within the Trust Region (Figure~\ref{fig_quant:1}).}
 (\textit{Left}) This scene shows a slow-moving construction vehicle, where the optimal trajectory is to follow carefully. The model's prediction aligns closely with the best-rated trajectory (Score 10.0), resulting in a perfect RFS of 10.0. (\textit{Center}) In this complex urban intersection, a cable car is moving while another vehicle is executing a right turn. The most preferred trajectory (Score 8.0) is to proceed carefully through the intersection, whereas the two lower-rated trajectories involve suboptimal actions like hard braking or deviating from the route. Since the model's prediction is well-aligned with the preferred path, it receives an RFS of 8.0. (\textit{Right}) The best behavior here is to safely nudge right to proceed past the bus without collision. The model's prediction accurately follows this optimal behavior, yielding an RFS of 10.0.

\noindent\textbf{Decayed Scores outside the Trust Region (Figure~\ref{fig_quant:2}).}
(\textit{Left}) In
snowy conditions, labeled trajectories include proceeding straight and turning left. The prediction follows the left-turn maneuver but at a slightly higher velocity than the labeled trajectory, causing the score to decay. (\textit{Center}) An oncoming motorcycle necessitates an avoidance maneuver. The prediction executes a similar lateral swerve at a comparable velocity but maintains a smaller lateral distance to the lane edge, resulting in a decayed score. (\textit{Right}) The objective is to proceed straight at a moderate velocity to avoid a cyclist approaching from the left. The prediction is significantly slower than the optimal (Score 10.0) trajectories, leading to a decayed score.

\noindent\textbf{Floor Scores for Predictions Far from Rater-Specified Trajectories (Figure~\ref{fig_quant:3}).}
(\textit{Left}) Labeled trajectories demonstrate both lane-following and a lane-change. The prediction, however, proceeds at a high velocity in the unrated region between the two maneuvers, thereby receiving the floor score. (\textit{Center}) While all labeled trajectories indicate a left turn, the prediction erroneously turns right. This significant deviation from the valid region results in the floor score. (\textit{Right}) The labeled trajectories execute a right turn. The prediction proceeds straight, diverging completely from the specified maneuvers and receiving the floor score.

\subsection{Benchmark Models}
Since the release of 
\ours, we have received a significant number of submissions utilizing various models. These can be broadly divided into three categories: MLLM-based, Diffusion-based, and MLP-based models. The following section details the methods that have released a detailed report, as shown in Table \ref{tab:main_table}.

\noindent\textbf{Swin-Trajectory~\cite{ParkSwinTrajectoryTR}} is a lightweight, MLP-based model. It uses a Swin Transformer~\cite{liu2021swin} to extract image features from three front cameras and a simple MLP to directly predict waypoints. The model is lightweight and achieves a slightly better RFS (7.543) than the baseline.

\noindent\textbf{DiffusionLTF and UniPlan} are both Diffusion-based models built on the DiffusionDrive~\cite{liao2025diffusiondrive} architecture. Their primary difference lies in the training datasets used: DiffusionLTF utilizes WOD-E2E, CARLA~\cite{dosovitskiy2017carla}, NAVSIM~\cite{Dauner2024NEURIPS}, and WOD-Perception~\cite{sun2020scalability}, whereas UniPlan is trained on WOD-E2E and nuPlan~\cite{caesar2021nuplan}. They achieve comparable performance, with RFS scores of 7.717 and 7.779, respectively.

\noindent\textbf{Poutine~\cite{pal2025poutinevisionlanguagetrajectorypretraining}, HMVLM~\cite{wang2025hmvlm}, and AutoVLA~\cite{zhou2025autovla}} are all MLLM-based models that use Qwen2.5 as their backbone. They share a similar problem formulation, taking camera images and ego states as input modalities and outputting future waypoints as text. Additionally, all three models utilize Chain-of-Thought (CoT) reasoning before generating a trajectory. Despite these similarities, their results show a significant performance gap, with AutoVLA achieving an RFS of 7.556, HMVLM 7.736, and Poutine 7.986. The primary differences among these methods stem from:

\begin{itemize}
\item  \textbf{Training data sources}: AutoVLA uses a combination of WOD-E2E, nuPlan, and nuScenes. In contrast, HMVLM is trained exclusively on WOD-E2E, whereas Poutine uses a blend of WOD-E2E and the CoVLA dataset.

\item \textbf{CoT captioning style} These three models employ different methods for generating reasoning captions and use distinct prompt templates.

 \item \textbf{RL training}: HMVLM does not include any post-training reinforcement learning. AutoVLA incorporates GPRO with ADE as the reward, whereas Poutine uses GPRO with RFS as the reward.

\end{itemize}

\subsection{Discussion of the Results}
From the results of these benchmark models, below we discuss important research questions in E2E Driving.

\noindent\textit{Q1: Is extra data source with large data distribution gap helpful for E2E Driving?}

\textbf{It depends}. For MLLM-based models, \eg Poutine and AutoVLA, adding extra data source is helpful, resulting in an obvious performance gain. However, for Diffusion-based models, \eg UniPlan and DiffusionLTF, only minor improvements are observed. A possible explanation for this divergence lies in the architectural capabilities of the MLLMs. We hypothesize that the CoT reasoning utilized by the MLLM-based models allows them to effectively leverage the diverse world knowledge and logical structures inherent in multiple datasets. This explicit reasoning mechanism helps the MLLMs internalize abstract driving knowledge that remains helpful regardless of the visual or geometric distribution shift between datasets. In contrast, diffusion-based models, which rely more directly on dense, pixel-level prediction, are more susceptible to performance degradation when combining visually disparate data sources.

\noindent\textit{Q2: Does a better ADE always lead to a better RFS?}

\textbf{No, a betterADE does not guarantee a better RFS.} We plotted a few data points from different model submissions, showing both their ADE and RFS scores in the right figure of Table~\ref{tab:main_table}. While the two metrics exhibit a rough positive correlation, we observe numerous models where better ADE performance does not translate to a higher RFS score. For instance, WayNet achieves a highly competitive ADE of 2.8, ranking among the best submissions, yet its RFS is significantly lower than most other models. Conversely, HMVLM demonstrates the opposite trend: its ADE is worse than many submissions, but its RFS ranks near the top. This clear divergence confirms the need for the RFS metric, as ADE alone is insufficient to evaluate a model's true effectiveness in handling safety-critical, multi-modal long-tail scenarios.

\noindent\textit{Q3: Is RL effective in E2E Driving?}

\textbf{Yes, particularly when the reward is aligned with the target evaluation metric.} Both Poutine and AutoVLA demonstrate performance improvements by incorporating RL into their post-training phase. However, the gain observed in Poutine is significantly more pronounced. The major reason for this difference lies in the reward signal used: Poutine utilizes RFS as its reward, which is directly aligned with our long-tail evaluation metric, whereas AutoVLA uses ADE. As the preceding research question demonstrated, ADE does not always maintain a strong positive correlation with RFS, making it a sub-optimal choice for optimizing performance on safety-critical scenarios.

\section{Conclusion}
\label{sec:conclusion}
In this paper, we introduced the Waymo Open Dataset for End-to-End Driving (WOD-E2E), a new benchmark specifically curated to evaluate end-to-end driving systems on challenging, long-tail scenarios. Existing datasets primarily feature nominal driving, failing to test true robustness. Our dataset provides 4,021 driving segments totaling approximately 12 hours, focusing on rare events that occur with a frequency of less than 0.03\%.

To overcome the limitations of traditional metrics like ADE in these complex, multi-modal situations, we also introducedsss a new metric: Rater Feedback Score (RFS). RFS is a novel, human-aligned metric that evaluates a model's trajectory against expert-annotated preference labels. Our benchmark analysis validates the dataset's utility, demonstrating a clear divergence between ADE and RFS scores. This confirms that RFS is essential for capturing true performance in safety-critical scenarios. The benchmark results also highlight the promise of MLLM-based models and the effectiveness of reinforcement learning when its reward is directly aligned with the RFS metric.

We adopted an open-loop setup for \ours\ due to the prohibitive computational cost of realistic sensor simulation. While this presents a limitation, \ours\ advances the state-of-the-art for open-loop E2E driving benchmarks. Moreover, the long tail real world driving scenarios in our dataset could be applicable for testing the generalizability of high-fidelity simulators. 

We hope that our WOD-E2E dataset and RFS metric will continue contributing to the development of more generalizable, robust, and safe autonomous driving agents.

\newpage
{
    \small
    \bibliographystyle{ieeenat_fullname}
    \bibliography{main}
}



\end{document}